\pdfoutput=1

\documentclass[11pt]{article}
\usepackage{authblk}
\usepackage{EMNLP2023}
\usepackage{mathtools}
\usepackage{cuted}
\usepackage{graphicx}
\usepackage{subfig}
\usepackage{times}
\usepackage{latexsym}
\usepackage{devanagari}
\usepackage{microtype}

\usepackage{algorithm,algpseudocode}
\usepackage{resizegather}
\usepackage{hyperref}
\usepackage{url}
\usepackage[T1]{fontenc}

\usepackage[utf8]{inputenc}

\title{Reference Free Domain Adaptation for Translation of Noisy Questions with Question Specific Rewards}

\makeatletter
\newcommand\email[2][]%
   {\newaffiltrue\let\AB@blk@and\AB@pand
      \if\relax#1\relax\def\AB@note{\AB@thenote}\else\def\AB@note{\relax}%
        \setcounter{Maxaffil}{0}\fi
      \begingroup
        \let\protect\@unexpandable@protect
        \def\thanks{\protect\thanks}\def\footnote{\protect\footnote}%
        \@temptokena=\expandafter{\AB@authors}%
        {\def\\{\protect\\\protect\Affilfont}\xdef\AB@temp{#2}}%
         \xdef\AB@authors{\the\@temptokena\AB@las\AB@au@str
         \protect\\[\affilsep]\protect\Affilfont\AB@temp}%
         \gdef\AB@las{}\gdef\AB@au@str{}%
        {\def\\{, \ignorespaces}\xdef\AB@temp{#2}}%
        \@temptokena=\expandafter{\AB@affillist}%
        \xdef\AB@affillist{\the\@temptokena \AB@affilsep
          \AB@affilnote{}\protect\Affilfont\AB@temp}%
      \endgroup
       \let\AB@affilsep\AB@affilsepx
}
\makeatother

\author[1]{\textbf{Baban Gain}}
\author[1]{\textbf{Ramakrishna Appicharla}}
\author[2]{\textbf{Soumya Chennabasavaraj}}
\author[2]{\\\textbf{Nikesh Garera}}
\author[1]{\textbf{Asif Ekbal}}
\author[2]{\textbf{Muthusamy Chelliah}}
\affil[1]{Department of Computer Science and Engineering, Indian Institute of Technology Patna, India}
\affil[2]{Flipkart, India}
\email{\{gainbaban, ramakrishnaappicharla, asif.ekbal\}@gmail.com}
\email{\{soumya.cb, nikesh.garera, muthusamy.c\}@flipkart.com}

\begin{document}
\maketitle
\begin{abstract}
Community Question-Answering (CQA) portals serve as a valuable tool for helping users within an organization. However, making them accessible to non-English-speaking users continues to be a challenge. Translating questions can broaden the community's reach, benefiting individuals with similar inquiries in various languages. Translating questions using Neural Machine Translation (NMT) poses more challenges, especially in noisy environments, where the grammatical correctness of the questions is not monitored. These questions may be phrased as statements by non-native speakers, with incorrect subject-verb order and sometimes even missing question marks.
Creating a synthetic parallel corpus from such data is also difficult due to its noisy nature. To address this issue, we propose a training methodology that fine-tunes the NMT system only using source-side data. Our approach balances adequacy and fluency by utilizing a loss function that combines BERTScore and Masked Language Model (MLM) Score. Our method surpasses the conventional Maximum Likelihood Estimation (MLE) based fine-tuning approach, which relies on synthetic target data, by achieving a 1.9 BLEU score improvement.
Our model exhibits robustness while we add noise to our baseline, and still achieve 1.1 BLEU improvement and large improvements on TER and BLEURT metrics. Our proposed methodology is model-agnostic and is only necessary during the training phase. We make the codes and datasets publicly available at \url{https://www.iitp.ac.in/~ai-nlp-ml/resources.html#DomainAdapt
} for facilitating further research.
\end{abstract}

\section{Introduction}
\label{sec:intro}
E-commerce decision-making heavily depends on community question-answering. When product descriptions and reviews fail to persuade users, they often turn to question-answering forums to address their concerns. However, English is used extensively on the majority of community question-answer portals. This situation renders it impossible for non-English speakers to ask questions and make informed purchasing decisions. Additionally, the potential loss of sales negatively affects businesses. Machine Translation (MT) is a valuable tool that enables users to communicate with individuals speaking different languages.

\begin{table}[]
\centering
\begin{tabular}{p{0.52\columnwidth}|p{0.38\columnwidth}}
\hline{}
\textbf{What User wants to say} & \textbf{What user wrote} \\ \hline
Does it work in Samsung A50S ?  & It works in samsung a50s \\ 
Will it fit in Xylo E4?         & In Xylo E4 will fit     \\\hline
\end{tabular}
\caption{An illustration of the mismatch between the input received by the MT system and its intended meaning. Considering that the sentence in the second column of the first row is intended to be a question, it is not grammatically correct.}
\label{tab:example-of-dataset}
\end{table}

Translating noisy questions differs significantly from general domain data, statements, or answers. Firstly, most currently available general domain data consists of statements rather than questions, making it less effective for translating questions. In the largest publicly accessible dataset \cite{ramesh-etal-2022-samanantar}, only 3.17\% of the total lines contain a question mark.
Secondly, questions exhibit a higher frequency of grammatical errors and are frequently presented as statements. For example, a sentence in the question field of a Community QA site - \textit{It works in samsung a50s}. In this case, the user intends to inquire about the product's compatibility with Samsung A50S, so the question should have been: \textit{Does it work in Samsung A50S ?}. The initial query is grammatically incorrect and appears more like a statement that the user knows the product works for the Samsung A50S. Moreover, the absence of a question mark (``?'') makes it difficult for both humans and automated systems to recognize it as a question unless they are aware it was posted in a community QA site's question field. In regions where English is not the first language, such grammatical errors in user queries are common, and this is particularly true of the Indian subcontinent, which has a very diverse linguistic population.

We aim to develop an NMT system that fluently translates English questions into a target language. In simpler terms, the input should be translated into the output, assuming grammatical correctness. This is challenging because the model must address grammatical errors that may seem grammatically correct at the sentence level. Furthermore, the manual creation of training sets for this data type is time-consuming and costly, as it involves annotating the intended input rather than just translating the text. Therefore, we avoid using parallel data\footnote{Technically, we supply the synthetic reference to the model. However, the references were used to sort the data according to their lengths to keep the training data order consistent among different models. The synthetic reference was not used to calculate the loss of the model.} for fine-tuning and instead utilize one pre-trained model as our baseline, fine-tuning it exclusively with source-side data.
This paper makes contributions in the following ways:
\begin{itemize}
    \item Our models deal with noisy data during training, which is very challenging in unsupervised domain-adaptation setting.
    \item Our method can translate sentences that appear grammatically correct on the surface but are grammatically incorrect when considering the contextual information that they are questions or queries.
    \item We propose a novel domain-adaptation method that balances adequacy and fluency without requiring references. 
    \item Existing unsupervised methods rely on target-side monolingual data, while our methods work on source-side monolingual data.
    \item Our models deal with noisy data during training, which is very challenging in unsupervised domain-adaptation settings. 
\end{itemize}
\section{Related Work}
Neural Machine Translation (NMT) has made significant progress in the past decade and has even reached human-level performance in certain domains and language pairs. However, research in the field of question-answering (QA) is still in its early stages, with only a few attempts having been made. 
\citet{vikram2018ambiguity-in-question-paper-translation,word-sense-ambiguity-in-questions} focused on translating academic question papers, primarily emphasizing word-sense disambiguation. It is important to note that the questions in this context were well-structured and grammatically correct. \citet{gain-etal-2022-low} tackled the translation of user-generated questions and enhanced the translation quality by incorporating answers alongside questions during the training process. This approach allowed the model to leverage contextual information from the answers. Furthermore, they employed fine-tuning techniques by training the model solely on questions or by using explicit \textit{question}/\textit{answer} tags to distinguish between them. Although these methods led to improvements in translation quality, fine-tuning questions alone produced similar results. However, it is important to highlight that the synthetic target-side dataset had limitations in addressing question-specific issues and using synthetic data posed risks related to hallucinations and grammatical errors. The creation of domain-specific noisy question annotations can be costly, rendering question translation infeasible using Maximum Likelihood Estimation (MLE) training. \citet{khayrallah-koehn-2018-impact} showed that training with noisy data can severely impact the results.  \citet{gain2023impact} explored the usage of visual context for the translation of noisy texts.
Alternative training methods, such as Minimum Risk Training (MRT) \cite{shen-etal-2016-minimum}, are employed to optimize model parameters with respect to arbitrary evaluation metrics, such as BLEU, to achieve superior translation outputs.  \citet{edunov-etal-2018-classical} observed that combinations of token-level and sequence-level losses outperformed the use of either loss type individually. It is worth noting that these methods also necessitate access to reference data, which makes them less suitable for translating noisy questions. As a result, the search for unsupervised methods becomes essential. 
\citet{wieting-etal-2019-beyond} introduced a loss function based on semantic similarity, which measures the similarity between hypotheses and references.
\citet{dou-etal-2019-unsupervised} harnessed target-side monolingual data to obtain domain-aware feature embeddings through language modeling tasks.
\citet{zheng-etal-2021-non-parametric} proposed the creation of a datastore for k-nearest-neighbor retrieval to facilitate domain adaptation in Neural Machine Translation (NMT) using target-side monolingual data. This method yielded comparable results to traditional back-translation techniques.
It is worth noting that all the unsupervised domain adaptation methods \citep{8546053, zheng-etal-2021-non-parametric} discussed in this context necessitate the availability of target-side monolingual data. However, we only have access to the source-side monolingual data. Therefore, we propose a novel method to fine-tune an NMT model using only source-side data, addressing the limitation of requiring target-side monolingual data in unsupervised domain adaptation, specifically in noisy text.


\section{Background}

The Neural Machine Translation (NMT) task can be divided into two major components: \textbf{Fluency}: Ensuring grammatical correctness in the generated output for the target language. \textbf{Adequacy}:  Preserving the meaning of the source text in the generated output.
NMT systems often produce translations that are fluent but may lack adequacy. \citet{voita-etal-2021-analyzing} suggested that this is partly because the models tend to prioritize partially translated output over the source sentences during the decoding stage. Achieving a balance between fluency and adequacy remains a challenging task.
Most NMT Systems use Maximum Likelihood Estimation (MLE) \cite{mle} objective during training. In the \autoref{eqn:maximum-likelihood}, where $S$ represents the number of training samples in a batch, $N$ is the number of tokens on the target-side of a training sample, $\mathbf{y}_n^{(s)}$ is the ground truth token on the target-side at step $n$, $\mathbf{x}^{(s)}$ is the source sentence, $\mathbf{y}_{<n}^{(s)}$ represents the target-side tokens from previous steps, and $\theta$ denotes the model parameters. Note that during training, the teacher forcing \cite{6795228-williams-teacher-forcing} method is used for faster convergence and stable training. In teacher-forcing method, ground truth tokens $\mathbf{y}_{<n}^{(s)}$ are used instead of partially translated output, $\mathbf{\hat{y}}_{<n}^{(s)}$. Major disadvantages of teacher forcing include
\begin{itemize}
    \item The trained model is exposed only to the training distribution but not its output.  However, the reference is not supplied to the model during testing. This makes the model completely rely on its (possibly wrong) partially translated output, creating a discrepancy between training and testing \cite{ranzato-exposure-bias}.
    \item Typically, evaluation metrics in NMT are applied at the sentence or document level. While the Maximum Likelihood Estimation (MLE) objective is effective in achieving high token accuracy, it may not yield optimal results for other metrics such as BLEU \cite{papineni-etal-2002-bleu}, TER \cite{snover-etal-2006-study}, COMET \cite{rei-etal-2020-comet}, etc. 
\end{itemize}

\begin{equation}
\mathcal{L}_{mle}(\theta)=\sum_{s=1}^S \sum_{n=1}^{N^{(s)}} - \log P\left(\mathbf{y}_n^{(s)} \mid \mathbf{x}^{(s)}, \mathbf{y}_{<n}^{(s)} ; \theta\right)
\label{eqn:maximum-likelihood}
\end{equation}
The challenges in NMT can be addressed through various approaches. For instance, a) gradually exposing the model to partially translated output as training progresses has been proposed as a solution \cite{zhang-etal-2019-bridging}. Alternatively, b) one can pre-train the model using MLE objectives before fine-tuning it with a desired evaluation metric, such as BLEU. These techniques have demonstrated their effectiveness in improving results up to a certain point.

Nonetheless, it is worth noting that the MLE objective often performs sufficiently well, especially when applied to clean, extensive datasets alongside suitable regularization techniques.
In real-world use cases, many organic datasets, including conversations, question-answers, and reviews, are primarily monolingual. While back-translation is recognized as an effective technique for leveraging monolingual data, it cannot be utilized when only source-side monolingual data is available.
Given that numerous datasets and websites are predominantly in English, and the construction of machine translation systems often involves translating from English to other languages, generating high-quality synthetic parallel datasets through forward-translation can be challenging. In back-translation, the source-side text is synthetic, while the target-side is considered the gold standard. Back-translation aids in robust training, as it introduces errors on the source-side due to the synthetic nature of the text, while the target-side remains correct. This is the opposite of forward-translation, making it somewhat less effective but valuable in situations where better alternatives are lacking in standard machine translation systems.

However, employing synthetic data through forward-translation can lead to adverse effects when dealing with noisy text translation. Because the source-side is inherently noisy, forward-translated synthetic data will inevitably contain a substantial amount of errors. This, in turn, results in the propagation of errors during model training.

Furthermore, using alternative evaluation metrics as loss functions is often not very helpful, as they rely on gold-standard references, such as the BLEU score, which may not be available. Therefore, we propose a novel loss function that relies solely on source-side sentences, a target-side language model, and a source-side grammatical error correction model.
\section{Methodology}
In this section, we first delve into MLM score and BERTScore and explain why we have chosen to incorporate them into our loss function. Subsequently, we detail our proposed training procedure for the translation of noisy questions.
\subsection{Masked Language Model Score}
\label{subsec:mlm-score}
The Language Model (LM) Score of a sentence can be described as in \autoref{eqn:lm-score}, where $y$ represents the sentence with $|y|$ tokens, and ${y}_{<n}$ denotes the tokens at previous positions in the sentence.
\begin{equation}
log P_{lm}(\mathbf{y})=\sum_{n=1}^{|\mathbf{y}|} \log P_{lm}\left(\mathbf{y}_n \mid  \mathbf{y}_{<n} \right)
\label{eqn:lm-score}
\end{equation}
\cite{salazar-etal-2020-masked-language-model-scoring-mlm} introduced a method known as Masked Language Model (MLM) Score. While the log-likelihood of a token in a Language Model (LM) is conditioned solely on previous tokens, in MLM, it is conditioned on both previous and next tokens, as described in \autoref{eqn:mlm-score}. Here, $\mathbf{y}_{\sim n}$ denotes all the tokens in the sentence except for the one at the $n$-th position. Notably, in contrast to LM, the MLM score does not suffer from a left-to-right bias.
\begin{equation}
log P_{mlm}(\mathbf{y})=\sum_{n=1}^{|\mathbf{y}|} \log P_{mlm}\left(\mathbf{y}_n \mid  \mathbf{y}_{\sim n} \right)
\label{eqn:mlm-score}
\end{equation}


\begin{table*}[]
\resizebox{\textwidth}{!}{%
\begin{tabular}{llll}
\hline
\textbf{Source}              & \textbf{Candidate}                & \textbf{Candidate (In English)} & \textbf{MLM Score} \\ \hline
It is suit for 1 Yr old baby & {\dn yh ek sAl k\? b\3CEw\? k\? Ele \8{s}V h\4.}  & It is suitable for 1 year old baby. & -1.46              \\
\multicolumn{1}{c}{User wanted to say:} & {\dn yh ek sAl k\? b\3CEw\? k\? Ele \8{s}V h\4{\rs ?\re}}      & It is suitable for 1 year old baby? & -2.58 \\
Does it suit for 1 yr old baby?            & {\dn \3C8wA yh ek sAl k\? b\3CEw\? k\? Ele \8{s}V h\4{\rs ?\re}}  & Is it suitable for 1 year old baby? & -1.51 \\ \hline
\end{tabular}%
}
\caption{Example illustrating the impact of different MLM scores when sentences are grammatically incorrect for a question. The MLM score of the first candidate is -1.46, but it is not appropriate for a question.  To address this, we removed the period ``{\dn .}'' and added a question mark ``?'' to the candidate in the second row, resulting in an MLM score of -2.58.  Thus, the model is encouraged to generate candidates that are appropriate for questions, as seen in the third row, to achieve a more favorable loss.}
\label{tab:mlm-score example}
\end{table*}
LM and MLM scores are commonly employed in MT re-ranking\cite{olteanu-etal-2006-language}. Typically, a set of $K$ candidate translations is generated using an NMT model. These candidates are then forwarded to either an LM or an MLM for scoring, and the candidate with the highest LM or MLM score is chosen as the final output for a given sentence. This re-ranking technique has been proven to be effective in improving results when compared to a basic model without re-ranking. Language Models tend to favor fluent sentences, which is advantageous in general-domain translation where most candidates are adequate.
However, in noisy scenarios, employing MLM scores for re-ranking is less effective. Firstly, candidate translations are more likely to be inadequate compared to non-noisy scenarios, so choosing the candidate with the best MLM score might lead to inadequacy. Furthermore, re-ranking functions as a pipeline between the NMT and MLM models, introducing additional processing time by passing candidate translations to the MLM. This results in increased testing time.
Hence, we employ the MLM score as the primary loss function during training to encourage the model to generate fluent utterances. For scoring the candidates, we utilize the \textit{bert-base-multilingual-uncased} model. Nonetheless, it is important to note that reinforcement learning (RL) models can exhibit reward-hacking behaviors when the rewards are not balanced. Since our rewards are currently based solely on fluency and not adequacy, the model may strive to produce highly fluent but contextually irrelevant sentences to achieve a better loss value.
To address this issue, we introduce BERTScore \cite{bertscore} to our loss function, aiming to strike a balance between adequacy and fluency. Further details about this approach are discussed in the following section.
\subsection{Pair-wise Cosine Similarity between Source and Candidate}
\label{subsec:bertscore}
As the target-side data is unavailable, widely used machine translation metrics like BLEU\cite{papineni-etal-2002-bleu}, METEOR\cite{banerjee-lavie-2005-meteor}, COMET, etc., which relies on human-annotated reference sentences cannot be used. Therefore, we search for metrics that can find similarities between the (noisy) source-side and the generated candidate on the target-side. It is essential for the metric to be multilingual to be able to handle the source and target side. Cosine similarity between the multilingual embeddings of the source and candidate appears to be a suitable choice for this purpose.
However, using contextual word embeddings \cite{10.5555/3295222.3295377-learned-in-translation, peters-etal-2018-deep, devlin-bert} is a more suitable approach, especially in noisy scenarios. Unlike traditional word embeddings, contextual word embeddings have the ability to capture word semantics from the context, allowing the word embedding to vary based on the context, even for the same word.
Once we obtain embeddings for source and candidate sentences, we calculate the recall as outlined in \autoref{eqn:bert-score-recall}. Essentially, this metric represents the sum of the highest similarity score of the most similar word in the candidate translation for each word in the source sentence. It is worth noting that since the vectors are pre-normalized,  calculation of $||\mathbf{x}_i||$ and  $||\hat{\mathbf{y}}_i||$ are not required in the cosine similarity formula.
\begin{equation}
R_{BERT}(\mathbf{\mathbf{x},\mathbf{\hat{y}}})=\frac{1}{|\mathbf{x}|} \sum_{\mathbf{x}_i \in \mathbf{x}} \max_{\hat{\mathbf{y}_j} \in \hat{\mathbf{y}}} 
\mathbf{x}_i^T \hat{\mathbf{y}_j}
\label{eqn:bert-score-recall}
\end{equation}
Similarly, precision is calculated as specified in \autoref{eqn:bert-score-precision}. In this context, precision reflects the sum of the highest similarity score of the most similar word in the source sentence for each word in the candidate translation.
\begin{equation}
P_{BERT}(\mathbf{\mathbf{x},\mathbf{\hat{y}}})=\frac{1}{|\mathbf{\hat{y}}|} \sum_{\mathbf{\hat{y}_j} \in \mathbf{\hat{y}}} \max_{{\mathbf{x}_i \in \mathbf{x}}} 
\mathbf{x}_i^T \hat{\mathbf{y}_j}
\label{eqn:bert-score-precision}
\end{equation}
Finally, F1 score is calculated with $R_{BERT}$ and $P_{BERT}$. 
\begin{equation}
    F_{BERT}=2 \times \frac{P_{BERT} \times R_{BERT}}{P_{BERT} + R_{BERT}}
\end{equation}
It is crucial to consider F1 instead of solely focusing on precision or recall. Relying only on precision might lead the models to generate very short sentences that are similar to some of the source words, making them fluent but not containing all the information from the source. Similarly, prioritizing only recall could encourage the model to generate longer sequences with most of the source words but also including words not present in the source.

For scoring our candidates, we employ the \textit{mbart-large-50-one-to-many-mmt} model \cite{tang2020multilingual}. In the subsequent sections, We will refer to BERTScore as $F_{BERT}$.

\subsection{Grammar Error Correction of Source}
\label{subsec:gec}
We've discussed how to address adequacy and fluency in the preceding sections. It's worth noting that adequacy is calculated using BERTScore between the source sentence and the candidate translation. However, due to the presence of noise in the source-side, this approach might penalize the NMT model when it attempts to generate robust candidate translations. For example, consider the source sentence \textit{What is the defference between the two}, where the spelling of \textit{different} is incorrect. If BERTScore is applied directly between the source and candidate, it will assign a lower similarity score if the NMT model produces a word similar to \textit{difference} instead of \textit{defference} in the target-side. Therefore, we use a publicly available Grammar Error Correction (GEC) model named Gramformer \footnote{\url{https://github.com/PrithivirajDamodaran/Gramformer}}. For an input sentence \textbf{x}, we obtain a grammatical correction version $\hat{\textbf{x}}$ using the Gramformer model. Nevertheless, it is important to note that the GEC model may occasionally make incorrect edits to the source sentence, resulting in a version that is worse than the noisy source itself. To ensure that we do not penalize the models for handling noise, we calculate the final similarity score as $max((BERTScore(\textbf{x},\hat{y}),(BERTScore(\hat{\textbf{x}},\hat{y}))$. Essentially, this approach considers the sentence with the highest similarity score as the most grammatically correct one.

\subsection{Proposed Model}
\begin{figure}
    \centering
    \includegraphics[width=0.9\columnwidth,keepaspectratio]{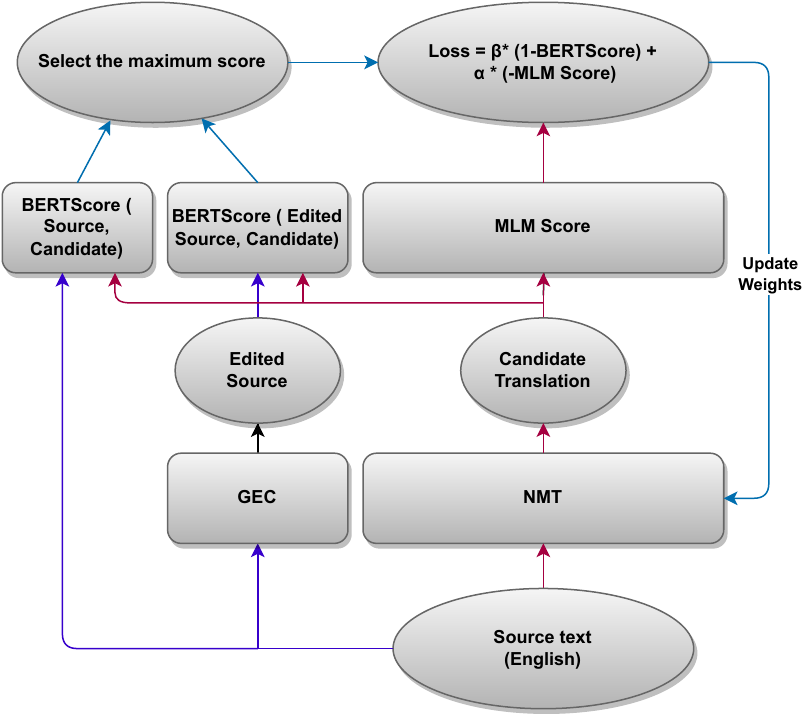}
    \caption{An abstract flow diagram of the training process}
    \label{fig:dataset_example}
\end{figure}

We use a combination of MLM (\autoref{subsec:mlm-score}), BERTScore (\autoref{subsec:bertscore}) and GEC model (\autoref{subsec:gec}) to train our model. First, we train an NMT model as a baseline (\autoref{subsec:baseline}) on general domain datasets.
We start by feeding the source sentence $\textbf{x}$ into the NMT model to generate $K$ candidate translations. To indicate that these sentences are questions, we append a question mark. However, it's important to note that this approach doesn't always work as intended; at times, the GEC model may interpret the presence of a question mark as a grammatical error and remove it from $\hat{\textbf{x}}$.
Subsequently, we calculate the similarity score between the source and each candidate using BERTScore. We repeat the same process with the edited source. For each candidate, we compute a metric called Similarity $\mathcal{L}_{BERT}$, which represents the maximum between the two BERTscores, subtracted from one, given that one is the maximum possible value for BERTScore.
Subsequently, we evaluate the fluency of the candidate translation using MLM. Finally, based on these scores, we calculate the loss and update the model parameters. It is important to emphasize that these scores serve the purpose of teaching the model to handle noise while maintaining fluency and adequacy. They are not required during testing. As a result, the architecture of the model remains unchanged."

\begin{algorithm*}
\label{algo:algorithm}

\begin{algorithmic}[1]

\State \textbf{MODEL} $\gets$ Pre-trained model; \textbf{GEC}  $\gets$ A source-side grammatical error correction model \;
\State \textbf{MLM}  $\gets$ A target-side/multilingual masked language model \;
\State index=0; K= Beam width for Minimum Risk Training
\For{input sentence $\textbf{x} \in batch$}
    \State $\hat{y}_{0:K}$ $\gets$ MODEL(\textbf{x})
    \State Remove full stops from the end, and if the last character of \textbf{x} or any of $\hat{y}_{0:K}$ is not ``?", append ``?". 
    \State $\hat{\textbf{x}}$ $\gets$ GEC(\textbf{x})
    \For{i=1 to K}
        \State $\mathcal{L}_{BERT} \gets 1 - \max(F_{BERT}(\textbf{x},\hat{y}_{i}), F_{BERT}({\hat{\textbf{x}},\hat{y}_{i}}))$
        \State $\mathcal{L}_{mlm} \gets - MLM(\hat{y}_{i})$
        \State Loss[index] = $P(\hat{y_i}| \textbf{x})$ $[\beta * \mathcal{L}_{BERT}+ \alpha * \mathcal{L}_{mlm}]$
        \State index = index + 1
    \EndFor    
\EndFor    
\State Total Loss = Sum of Losses at each index
\State Repeat Steps 3-15 for the designated number of training steps.

 \caption{Our Proposed Training Procedure}
\end{algorithmic}
\end{algorithm*}

\begin{figure*}
\begin{equation}
    \mathcal{L}_{rl}(\theta)=\sum_{s=1}^S \left(\sum_{\hat{y} \in \mathcal{Y}(x^{s})} \left[\sum_{n=1}^{N^{(s)}} \log P\left(\mathbf{\hat{y}}_n^{(s)} \mid \mathbf{x}^{(s)}, \mathbf{\hat{y}}_{<n}^{(s)} ; \theta\right)\right] \cdot \mathcal{F}(\mathbf{x}^{(s)},\mathbf{\hat{y}}^{(s)}) \right)
\label{eqn:final-loss}
\end{equation}
\end{figure*}

The final loss function is presented in \autoref{eqn:final-loss}. In this function, we multiply the probability of the generated sequence with $(\mathcal{F}(\mathbf{x}^{(s)},\mathbf{\hat{y}}^{(s)})$. Here, $S$ represents the number of sentences in a batch, and ${N^{(s)}}$ denotes the number of tokens in the generated candidate. Due to the exponential search space of $\mathcal{Y}(x^{s})$, we employ a sampling approach with K=5 candidates per training sentence, where $y \in \mathcal{Y}(x^{s})$. Beam search is utilized to prevent the duplication of candidates.
\begin{gather}
    \mathcal{F}(\mathbf{x}^{(s)},\mathbf{\hat{y}}^{(s)}) = \alpha \cdot \mathcal{L}_{MLM}(\mathbf{\hat{y}}^{(s)})  + \beta \cdot  \mathcal{L}_{BERT}(\mathbf{x}^{(s)},\mathbf{\hat{y}}^{(s)})
\end{gather}
$\mathcal{F}(\mathbf{x}^{(s)},\mathbf{\hat{y}}^{(s})$ is determined as a weighted average between BERTScore Loss and MLM Score Loss. Notably, the MLM score tends to exhibit higher variance in comparison to BERTScore. To optimize the model's performance, we experimented with different sets of weights for $\alpha$ and $\beta$ and discovered that the weights 0.15 and 0.85 yielded the best results. From \autoref{eqn:mlm-score}, we find $\mathcal{L}_{MLM} = - log P_{mlm}(\hat{y})$.
\begin{gather}
\mathcal{L}_{BERT}(\mathbf{x}^{(s)},\mathbf{\hat{y}}^{(s)})=1 - \left(max((F_{BERT}(\textbf{x},\hat{y}),(F_{BERT}(GEC({\textbf{x}}),\hat{y}))\right)
\label{eqn:berscore-loss}
\end{gather}
Finally, we compute $\mathcal{L}_{BERT}$ by subtracting the maximum value of $F_{BERT}$ from one, given that one represents the highest possible value for $F_{BERT}$.
\section{Experiments and Results}
In this section, we will provide an overview of the datasets used, discuss the baseline models, and present the results achieved by our model.
\subsection{Dataset and Annotation}
For pre-training our NMT model, we utilize the Samanantar corpus, which comprises over 10 million sentence pairs for English-Hindi in the general domain. During the fine-tuning process, we focus on the first 50,000 questions from the Flipkart QnA corpus \cite{gain-etal-2022-low}, using only the English side of the data. In our evaluation, we employ both sides of the test set, consisting of 500 questions. It is important to note that we manually edited some of the references in the test set to enhance fluency, maintain consistency, and ensure user-friendliness. These edits took into account product names, types, and other relevant details to make the references more suitable for questions.

Additionally, we use the Mintaka dataset \cite{sen-etal-2022-mintaka} for evaluating our methods. Although Mintaka is a multilingual question-answering dataset, we repurpose it for translation tasks in this study. Notably, during training, we deliberately excluded the Hindi and German sides of the dataset to simulate a scenario in which the target-side of the training data is unavailable."
\begin{table}[]
\resizebox{\columnwidth}{!}{%
\begin{tabular}{llll}
\hline
\textbf{Model}        & \textbf{BLEU}        & \textbf{TER}        & \textbf{BLEURT} \\ \hline
\textbf{Baseline}     & 43.8        & 39.4    &  0.7507   \\ 
\textbf{+ MLE finetune} & 45.3        & 38.0    &  0.7613   \\
\textbf{+ GUDA} &     43.3    &  40.2   &  0.7470   \\
\textbf{+ MRT with BLEU } & 45.5 (+0.2)       & 38.5 (+0.5)   & 0.7628    \\
\textbf{+ Ours}         & 47.2 (+1.9) & 35.8 (-2.2) & 0.7646  \\ \hline
\textbf{Baseline + Noise}     & 44.9        & 39.1    &  0.7602   \\ 
\textbf{+ MLE finetune} &     45.7    &  37.8   &  0.7601   \\
\textbf{+ GUDA} &     43.1    &  40.5   &  0.7480   \\
\textbf{+ MRT with BLEU } &    46.3 (+0.6)   &   37.6 (-0.2)  &  0.7600   \\
\textbf{+ Ours}         & 46.8 (+1.1) & 35.4 (-2.4) &  0.7742 \\ \hline
\end{tabular}%
}
\caption{Results of our method on Flipkart QnA corpus (En-Hi) }
\label{tab:results-questions}
\end{table}
\subsection{Baseline}
\label{subsec:baseline}
We obtain a pre-trained model from \cite{gain-etal-2022-low}, which was trained on a large-scale English-Hindi data and use it as our baseline. The baseline consists of standard transformer architecture with six encoders and six decoders. The model is trained on 10.9 million general-domain English-Hindi sentence pairs obtained from \cite{ramesh-etal-2022-samanantar}. The model achieves 43.8 BLEU,
39.4 TER and 0.7507 BLEURT \cite{sellam-etal-2020-bleurt} scores, respectively.
\subsection{Robust Baseline}
Since we deal with noisy user-generated content, for robust training, we implement the following:
We apply three types of noise on the source-side of the pre-training dataset. a) \textbf{Natural Noise:} We replace characters with random characters with 1\% probability.
b) \textbf{Keyboard Noise:} We replace characters with surrounding characters from the keyboard with 5\% probability.
c) \textbf{Vowel Removal:} Users often do not type vowels. We drop vowels with 5\% probability.
Then, we combine the clean data with noisy data and train the model.
 We obtain 44.9 BLEU score with robust baseline, which is a +1.1 improvement from the non-robust baseline.

\subsection{Domain Adaptation with MLE}
We generate synthetic target data by translating the in-domain datasets with the baseline model. 
Similarly, we generate synthetic target data from the robust baseline.
Then, we initialize the model's weights from the baseline model and fine-tune the model on the respective synthetic data.
Note that we did not add noise at this stage as the in-domain dataset is already noisy.
We use label-smoothed cross entropy as the loss function and set 0.1 as the smoothing value. The model achieves 45.3 BLEU, 38.0 TER, and 0.7613 BLEURT scores. It outperforms the baseline model by 1.5 BLEU and 1.4 TER scores. 
After fine-tuning with the robust baseline, we achieve BLEU and TER scores of 45.7 and 37.8, respectively. This indicates a +0.4 BLEU improvement over the MLE method without robust pre-training.

\subsection{Unsupervised Domain Adaptation with Cross-Lingual Data Selection}
\citet{vu-etal-2021-generalised} proposed a generalized unsupervised domain adaptation technique (GUDA) for NMT where only monolingual data from either the source or target language is available in the new domain. A cross-lingual data selection method is introduced to select relevant in-domain sentences from a large monolingual corpus for the language without in-domain data. This involves learning an adaptive layer on top of multilingual BERT using contrastive learning to align source and target language representations. A domain classifier trained on the available in-domain monolingual data can then be transferred cross-lingually to select relevant data in the other language. We sample 500K sentence pairs from general domain data and select 50K sentence pairs from the sampled dataset with cross-lingual data selection. However, the selected examples are mostly noisy and not relevant to the target domain (here, noisy questions). This can be attributed to the fact that our target domain dataset contains noise, resulting in improper data selection. Consequently, this method deteriorated the results compared to baselines. As training progresses, we observe a drop in validation set results. This decline occurs because the model's performance degrades over longer training with noisy data.

\subsection{Domain Adaptation with BLEU}
Similar to the MLE method, we generate synthetic references from baseline and robust baseline models.
We use 1-BLEU as the loss function and train with the MRT method. We achieve 1.7 BLEU and 0.9 TER improvements over the baseline. However, it is important to note that the improvements vary across different metrics when comparing this method to MLE.
Using the robust baseline helped to achieve superior performance due to improved synthetic references. We achieve 0.6 and 0.2 BLEU and TER improvement, while BLEURT remained the same. Note that improvements achieved by this method w.r.t. the corresponding MLE based methods are statistically insignificant \cite{koehn-2004-statistical}, which can be attributed to the noisy nature of synthetic references.

\subsection{BERTScore and MLM Loss}
\begin{table*}[]
\centering

\begin{tabular}{lll}
\hline
\textbf{Model} & \textbf{Translation}               & \textbf{(In English)} \\ \hline
\multicolumn{3}{c}{Source: Does is Support on Hyundai i10.}                             \\ \hline
\textbf{Baseline / MLE / BLEU }       & {\dn yh \7{h}\2X\4i }i{\dn \rn{10} ko spoV\0 krtA h\4.}     & It supports Hyundai i10.          \\
\textbf{Ours}           & {\dn \3C8wA \7{h}\2X\4i aAI\rn{10} ko spoV\0 krtA h\4{\rs ?\re}}  & Does (it) support Hyundai i10?   \\ \hline
\end{tabular}%
\caption{An Example of translation generated by our NMT models}
\label{tab:manual-example}
\end{table*}
We report the results of our proposed method in \autoref{tab:results-questions} and \autoref{tab:results-questions-added-question}. We achieve 47.2 BLEU, 35.8 TER, and 0.7646 BLEURT scores. Our method outperforms the baseline model by 3.4 BLEU points and improves TER by 3.6 points. Further, this method outperforms the MLE model trained with synthetic data by 1.9 BLEU score and 2.2 TER points and MRT with BLEU as the loss function by similar margins.
Upon using a robust model as a baseline, we achieve 46.8 and 35.4 TER scores, which is 1.1 BLEU and 2.4 TER improvements compared to the MLE method. 
We perform a statistical significance test between the outputs of this method and the outputs of the corresponding MLE method with the Moses Toolkit \cite{koehn-2004-statistical,koehn-etal-2007-moses}. We found that the improvements are statistically significant, with p-values of 0.002 and 0.03 for non-robust and robust models, respectively, with respect to the corresponding MLE fine-tuned models.
It is worth noting that when using a robust baseline for our proposed method, the BLEU score decreases by a small margin (-0.4 BLEU). In contrast, when using MRT with the MLE method, the BLEU score increases by 0.4. This suggests that robust pre-training has a limited effect when fine-tuning is performed on noisy data.
\subsection{Results on Mintaka dataset}
\begin{table}[]
\resizebox{\columnwidth}{!}{%
\begin{tabular}{llll}
\hline
\textbf{Model}        & \textbf{BLEU}        & \textbf{TER}        & \textbf{BLEURT} \\ \hline
\multicolumn{4}{c}{\textbf{English-Hindi}} \\
\textbf{Baseline}     & 26.6        & 57.4    &  0.7335   \\ 
\textbf{Ours} & 27.6   (+1.0)     & 55.5 (-1.9)   &  0.7371 \\
\multicolumn{4}{c}{\textbf{English-German}} \\
\textbf{Baseline}     & 47.5        & 38.4    &   0.8334  \\ 
\textbf{Ours} & 47.8  (+0.3)       & 38.1 (-0.3)    &  0.8344 \\ \hline
\end{tabular}%
}
\caption{Results of our method on Mintaka dataset }
\label{tab:results-mintaka}
\end{table}
To simulate a scenario where only source data is available, we refrain from using the target-side of the data during training. Consequently, we compare the models with pre-trained models. Given that the questions are typically grammatical and the question mark is present in the source, we do not need to add it explicitly. Our method leads to a notable improvement of 1.0 BLEU points for English-Hindi compared to the baseline. However, the improvement is more modest, with just 0.3 BLEU points for English-German. Both English and German are considered high-resource languages, and the baseline model is trained on a large dataset. Therefore, the baseline model can accurately translate most of the questions, given that the sentences in the Mintaka dataset are non-noisy. This limits the potential for improvement over a strong baseline when a parallel corpus is unavailable.

\subsection{Analysis}
We have observed that robust pre-training significantly improves our results. Nevertheless, the degree of improvement diminishes after fine-tuning, as both robust and non-robust baselines are fine-tuned with noisy data and learn to handle noise to a similar extent.
We manually inspect sentences and check how our proposed method improves the performance.
We provide one example in \autoref{tab:manual-example}. Note that the source sentence is grammatically incorrect. First, the sentence contains \textit{is} instead of \textit{it}. Further, it contains a full stop instead of a question mark. Baseline and MLE models were unable to handle it. However, our method was able to generate the correct translation. Note that although our model generated correct outputs in many such instances, there exists a large number of samples where the model was unable to generate question-like translation. We observed that, even though our model was able to increase the probability of question-like candidates, often it is still lower than statement-like candidates. We suggest that this is due to MLE pre-training. We would like to explore to avoid this in our future work.

\subsection{Limitation}
This method should be preferred when there is very little or no high-quality parallel corpus available for domain adaptation. In a non-noisy situation, it might be more effective to use a robust model to generate synthetic data. Note that the proposed loss function has high variance due to the presence of MLM score, and checkpoints should be frequently saved to get the optimal results. The loss functions rely on BERTScore and MLM models, which are known to be subject to biases \cite{sun-etal-2022-bertscore,jentzsch-turan-2022-gender,zhang-hashimoto-2021-inductive} that can propagate to the NMT model. While we did not observe such instances in our limited studies, it is important to remain vigilant about potential biases. It is essential to exercise caution when applying this method in domains where a mistranslation could have severe consequences, such as medical question-answering portals. We believe that the general concept presented in this paper may have relevance for other generative tasks that require balancing different aspects of the outputs. While exploring this is beyond the scope of our current work, it is a direction we plan to investigate in the future.
\section{Conclusion}
We have developed a robust NMT system tailored for translating questions. Our focus is on addressing the unique challenges posed by noisy questions, which are often presented in the form of statements due to the limited grammatical knowledge of users.
The MLE-based fine-tuning with synthetic data has several limitations, specifically when the source is noisy. We propose an MLM and BERTScore-based training method to balance adequacy and fluency instead of using synthetic references for training data. Our method improves translations for noisy questions compared to MLE fine-tuning with synthetic data, and it also enhances translations on non-noisy data compared to the pre-trained model. We achieve up to 47.2 BLEU and 35.4 TER scores, based on different settings. We conducted human evaluations with annotators from an E-commerce organization and observed a clear improvement in translation quality. We believe that the approach of balancing fluency and adequacy during training can be applied to other domains and languages. In the future, we plan to explore the use of Quality Estimation metrics, capable of scoring both fluency and adequacy during training. Further, we would like to explore extending the method for other low-resource languages.
\section{Ethical Declaration}
We have used publicly available datasets and content from CQA portals for training purposes, ensuring compliance with copyright regulations. To our knowledge, no personal information has been utilized in our training data. It is important to note that while our procedure has potential benefits, it is not entirely foolproof and should be used with moderation, and we have highlighted its limitations in our paper.
\section{Acknowledgement}
Authors gratefully acknowledge the unrestricted research grant received from the Flipkart Internet
Private Limited to carry out the research. 

\bibliography{anthology,custom}
\bibliographystyle{acl_natbib}
\clearpage
\appendix
\section{Experimental Setup}
We use the fairseq \cite{fairseq} library for our implementation. For English-German, we use a publicly available baseline from \citet{ng-etal-2019-facebook}. We use standard transformer \cite{NIPS2017_3f5ee243-attention-is-all-you-need} architecture for English-Hindi and transformer-large for English-German.  we use 0.2 as the dropout \cite{JMLR:v15:srivastava14a-dropout} value for all our experiments. We set the maximum source tokens per batch of training to 200 and continue training for 5000 steps. We save checkpoints every 250 steps and use Adam \cite{adam-optimizer} optimizer with $\beta_{1} = 0.9$ and $\beta_{2} = 0.98$. We use the other default hyper-parameters from \cite{NIPS2017_3f5ee243-attention-is-all-you-need} and select the checkpoint with the best BLEU score for all the models. It takes approximately 15 minutes to train MLE models and about 10 hours to train our proposed method for 5000 steps. However, we believe there is a scope for better parallelization on our method, although the training time will still be higher due to beam search during training and tokenization during the calculation of BERTScore and MLM score. For experiments on Mintaka dataset, we use the same baseline that we employed for the Flipkart Questions dataset (\autoref{subsec:baseline}) for English-Hindi direction. In the case of English-German, we utilize Meta AI's WMT19 pre-trained translation model \cite{ng-etal-2019-facebook}. We use NVIDIA A100-PCIE-40GB GPU for training our models. We use sacrebleu \cite{post-2018-call} to calculate BLEU and TER scores. Our English-to-German model is made up of transformer-large that contains 313M parameters, while the English-to-Hindi model contains 71M parameters with transformer-base architecture. For MLM scoring, we use \textit{bert-base-multilingual-uncased}, which contains 110M parameters. Gramformer is a T5-based model containing 220M parameters. For BERTScore, we use \textit{facebook/mbart-large-50-one-to-many-mmt} with 610M parameters. Note that these weights are non-trainable and are not used during the testing phase. Samanantar corpus is released under \textit{CC BY-NC 4.0} License. English-to-German model is pre-trained on multiple datasets \cite{ng-etal-2019-facebook} and Mintaka dataset containing 14K train and 4K valid sentences is released under CC BY 4.0 License. We use Moses Toolkit \cite{koehn-etal-2007-moses} for pre-processing and fastBPE \cite{sennrich-etal-2016-neural} for byte-pair-encoding. The codes and datasets can also be accessed from \url{https://github.com/babangain/unsup_questions_translation}.

\section{Human Evaluation}

We obtain real-world test sets from a well-known e-commerce organization and translate them with our models. We request the organization to evaluate the translations manually for Quality Control.
We report the results of the manual evaluation on \autoref{tab:human-eval}. \textit{Good} denotes that the generated translation is of excellent quality and requires no further adjustments. \textit{Can be Better} indicates some issues in the translation, such as punctuation errors, poor word choices, transliterated words instead of translations, or vice versa, and other minor concerns. \textit{Bad} indicates that there is a mismatch between the candidate and the actual meaning of the source, or the candidate is not fully adequate, etc. The MLE-based model is able to achieve 9.4\% Good and 62.2\% Can be Better ratings, which is an improvement of 1.6\% and 2.2\% from baseline in absolute terms. With our proposed methodology, we achieve 9.4\% of Good and 64\% of Can be Better ratings, which is 2\% and 4\% improvement from baseline in absolute terms. Note that our method achieves better ratings compared to MLE fine-tuning with synthetic target data, even though our method does not use any target dataset.

\begin{table}
\resizebox{\columnwidth}{!}{%
\begin{tabular}{llll}
\hline
\textbf{Rating}        & \textbf{Baseline} & \textbf{MLE}  & \textbf{Ours} \\ \hline
\textbf{Good}          & 7.8      & 9.4  & 9.8  \\
\textbf{Can be Better} & 60       & 62.2 & 64   \\
\textbf{Bad}           & 32.2     & 28.4 & 26.2 \\ \hline
\end{tabular}%
}
\caption{Results of our methods on human evaluation (in percentage)}
\label{tab:human-eval}
\end{table}
\begin{table}[]
\resizebox{\columnwidth}{!}{%
\begin{tabular}{llll}
\hline
\textbf{Error Type}                       & \textbf{Baseline} & \textbf{MLE} & \textbf{Ours} \\ \hline
\textbf{Mismatch}            & 5.2               & 4.4          & 2.8           \\
\textbf{Words Missed}        & 12.0              & 8.4          & 5.0           \\
\textbf{Bad choice of words} & 21.8              & 23.8         & 26.0 (3.6) \\
\textbf{acronym/abbr transliterated} & 6.2              & 6.8         & 7.4 (1.4)\\ \hline
\end{tabular}%
}
\caption{Category-wise errors of our methods on human evaluation (in percentage); The number within the bracket indicates the percentage of errors of that category where the error was \textit{Mismatch} or \textit{Words Missed} with the baseline model. }
\label{tab:human-eval-category}
\end{table}

We present a categorical error report in \autoref{tab:human-eval-category}. Note that only one type of error is chosen per translation. If a sentence contains multiple issues like \textit{Words missed} and \textit{Bad choice of words}, the more serious issue (here \textit{Words missed})  is chosen as the error type.  We observe that serious issues like \textit{Mismatch} and \textit{Words Missed} are reduced with our proposed models. Minor errors, such as \textit{Bad choice of words} and \textit{acronym/abbreviation got transliterated} increased because the sentences that were producing critical errors with other models are now producing minor errors with our proposed model. For instance, even though \textit{Bad choice of words} increased by 4.2\%, most of them (3.6\%) come from the model's ability to generate better quality translation. Further, \textit{acronym/abbreviation got transliterated} increased by 1.2\% with our model, but 1.4\% of them comes due to superior quality translation with our method. Therefore, the model reduced 0.2\% of the error in other cases. While there exist multiple error categories, we reported the results in the categories where there is a major change in the numbers with the used models.

\section{Adding Question Marks during Testing}

\begin{table}[H]
\resizebox{\columnwidth}{!}{%
\begin{tabular}{llll}
\hline
\textbf{Model}        & \textbf{BLEU}        & \textbf{TER}        & \textbf{BLEURT} \\ \hline
\textbf{Baseline}     & 46.6        & 37.8    &   0.7696  \\ 
\textbf{MLE finetune} & 48.9        & 35.3    &   0.7830 \\
\textbf{Ours}         & 50.2 (+1.3) & 33.5 (-1.8) & 0.7864  \\ \hline
\end{tabular}%
}
\caption{Results of our method on questions dataset after adding question marks when absent, to assist the models in interpreting the input as a question during Testing on Flipkart QnA corpus (En-Hi) }
\label{tab:results-questions-added-question}
\end{table}
 We also append question marks with the input sentences before forwarding them to the NMT model. For the baseline, We achieve 46.6 BLEU, 37.8 TER, and 0.7696 BLEURT scores. Similarly, we achieve 48.9 BLEU, 35.3 TER, and 0.7830 BLEURT scores with the MLE method. This outperforms the baseline model with added question marks by 2.3 BLEU and 2.5 TER scores.
 We achieved 50.2 BLEU, 33.5 TER, and 0.7864 BLEURT scores with our proposed method. Under this setting, our method outperforms the baseline model by 3.6 BLEU and 4.3 TER scores. Further, it outperforms the MLE model with synthetic data by 1.3 BLEU and 1.8 TER scores.
 \section{Choice of Evaluation Metrics}
 We use BLEU and TER as these two are the most popular metrics, which often but not always correlate with human judgment. Recent metrics like COMET \cite{bosselut-etal-2019-comet} and COMET-QE have shown very promising co-relations with human judgment. However, the COMET metric is based on source, hypotheses, and reference, while COMET-QE is based on source and hypotheses. Since the source is noisy, deep learning based metric, which depends upon token embedding, will not be able to generate faithful results due to noisy embeddings. Therefore we use BLEURT as the third metric for evaluation since it depends upon hypotheses and reference, and references for the test set is manually created by human annotators. Note that BLEURT has the highest Kendall $\tau$ compared to competing metrics \cite{lee2023survey} like BARTScore \cite{NEURIPS2021_e4d2b6e6-bartscore} or BEER \cite{stanojevic-simaan-2014-fitting}. 
 \begin{figure*}[btp]%
    \centering
    \subfloat[\centering Top 5 candidates with MLE Model]{{\includegraphics[width=7cm]{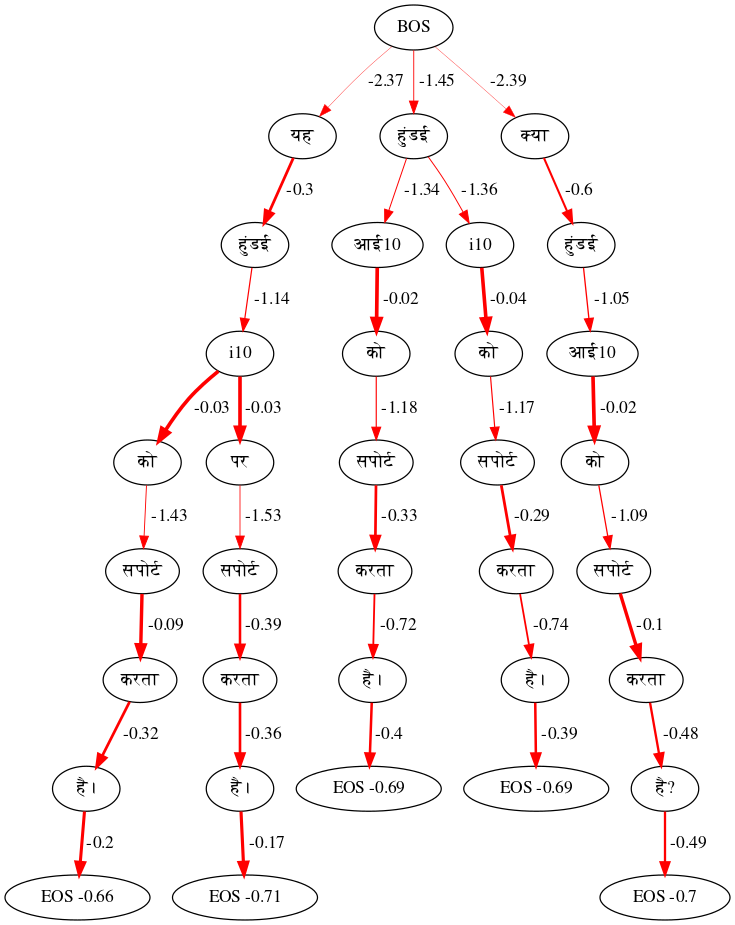} }}%
    \qquad
    \subfloat[\centering Top 5 candidates with Our Model]{{\includegraphics[width=7cm]{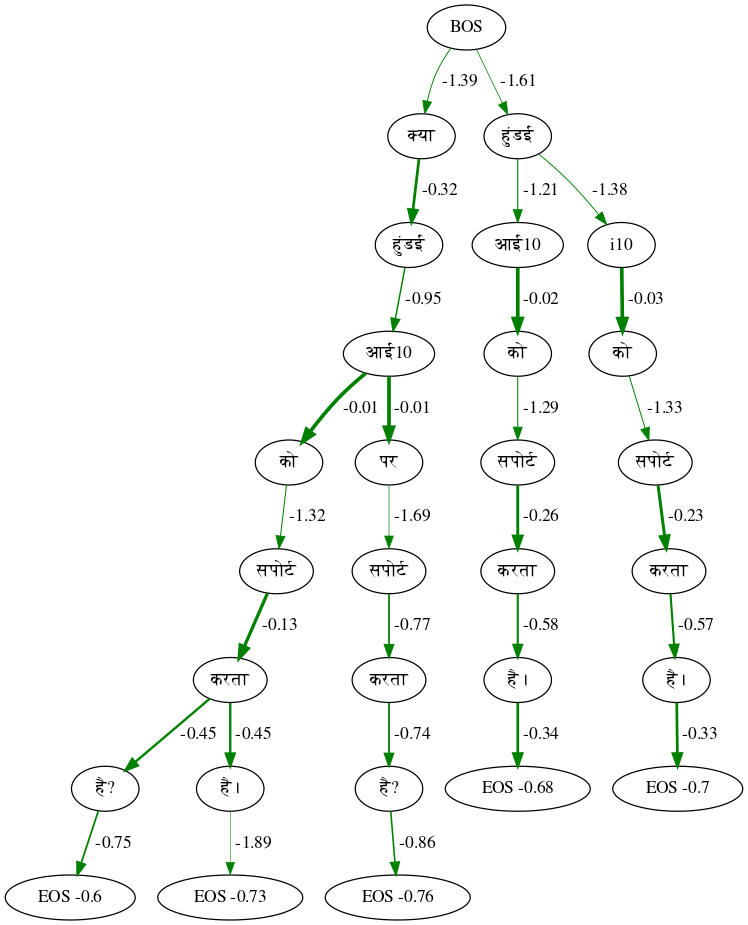} }}%
    \caption{Beam search tree of \textit{does is support on hyundai i10.} on the models with beam width=5}%
    \label{fig:example-beam}%
\end{figure*}
 \section{Sensitivity of $\alpha$ and $\beta$}
 We obtain the best results when $\alpha$=0.15 and $\beta$=0.85. We tried with different values like (0.5,0.5), (0.4,0.6), (0.3,0.7), (0.2,0.8), (0.1,0.9), (0.5,1.0). For $\alpha$=0.2 and $\beta$=0.8, we obtain 45.5 BLEU. Similarly,  We obtain 45.4 BLEU with $\alpha$ and $\beta$ values of 0.1 and 0.9, respectively. In contrast, we obtain 47.2 BLEU with 0.15 and 0.85 as $\alpha$ and $\beta$ values.
Increasing $\alpha$ too much resulted in fluent but non-adequate translation, whereas increasing $\beta$ resulted in adequate but non-fluent translation.

\section{Using GEC during Inference}
We also try to observe if the GEC model can be used during testing to achieve even better results. Firstly, we add question marks to the end of sentences where it is absent. Then, we pass it to the GEC model, and finally, we pass its output to the NMT model. In \autoref{tab:results-questions-added-question-then-grammar}, we report our results and observe mixed improvement in the results. Note that the difference in the bracket indicates the improvement when GEC is not used during testing time. Since the GEC model is not exclusively trained on questions, it tends to remove question marks from sentences to make them more like statements. We suggest that training a GEC model exclusively on questions could improve the results. However, it is difficult to train a high-quality GEC model for questions since the size of question datasets is much lower compared to general-domain data. 
\begin{table}[H]
\resizebox{\columnwidth}{!}{%
\begin{tabular}{llll}
\hline
\textbf{Model}        & \textbf{BLEU}        & \textbf{TER}        & \textbf{BLEURT} \\ \hline
\textbf{Baseline}     & 46.7 (+0.1)       & 37.3 (-0.5)   &   0.7812  \\ 
\textbf{MLE finetune} & 48.2  (-0.7)      & 35.6 (+0.3)   & 0.7907   \\
\textbf{Ours}         & 49.6 (-0.6) & 33.4 (-0.1) & 0.7949  \\ \hline
\end{tabular}%
}
\caption{Results of our method on questions dataset after adding question marks and then passing to GEC model during Testing}
\label{tab:results-questions-added-question-then-grammar}
\end{table}

\section{Example of Candidates with Beam Search During Testing}

In \autoref{fig:example-beam}, we show the beam search tree of two of our models. The numbers in the EOS nodes indicate the log-likelihood of the path. Note that with MLE, most of the candidates are like statements. Only one candidate appears in the top 5 (last branch), which ranks fourth among the top five candidates. However, with our method, the question-like candidate has a much higher probability compared to other sentences. The top (first branch) and fifth candidate (third branch) are question-like, and the fourth one (second branch) is partially like a question.

\end{document}